\documentclass[sigconf]{acmart}

\AtBeginDocument{%
  \providecommand\BibTeX{{%
    \normalfont B\kern-0.5em{\scshape i\kern-0.25em b}\kern-0.8em\TeX}}}

\copyrightyear{2023}
\acmYear{2023}
\setcopyright{rightsretained}
\acmConference[ICMI '23 Companion]{INTERNATIONAL CONFERENCE ON MULTIMODAL INTERACTION}{October 9--13, 2023}{Paris, France}
\acmBooktitle{INTERNATIONAL CONFERENCE ON MULTIMODAL INTERACTION (ICMI '23 Companion), October 9--13, 2023, Paris, France}\acmDOI{10.1145/3610661.3617151}
\acmISBN{979-8-4007-0321-8/23/10}

\usepackage{multirow}
\usepackage{colortbl,array,xcolor}
\usepackage{multicol}
\usepackage{graphicx}
\usepackage{pbalance}

\newcommand{\vsapceundergifure}{\vspace{0mm}}

\begin{document}

\title[Towards Objective Evaluation of Socially-Situated Conversational Robots]{Towards Objective Evaluation of\\Socially-Situated Conversational Robots:\\Assessing Human-Likeness through Multimodal User Behaviors}

\author{Koji Inoue}
\email{inoue@sap.ist.i.kyoto-u.ac.jp}
\affiliation{%
  \institution{Kyoto University}
  \state{Kyoto}
  \country{Japan}
}

\author{Divesh Lala}
\email{lala@sap.ist.i.kyoto-u.ac.jp}
\affiliation{%
  \institution{Kyoto University}
  \state{Kyoto}
  \country{Japan}
}

\author{Keiko Ochi}
\email{ochi@sap.ist.i.kyoto-u.ac.jp}
\affiliation{%
  \institution{Kyoto University}
  \state{Kyoto}
  \country{Japan}
}

\author{Tatsuya Kawahara}
\email{kawahara@i.kyoto-u.ac.jp}
\affiliation{%
  \institution{Kyoto University}
  \state{Kyoto}
  \country{Japan}
}

\author{Gabriel Skantze}
\email{skantze@kth.se}
\affiliation{%
  \institution{KTH Royal Institute of Technology}
  \state{Stockholm}
  \country{Sweden}
}

\renewcommand{\shortauthors}{Inoue, et al.}

\begin{abstract}
This paper tackles the challenging task of evaluating socially situated conversational robots and presents a novel objective evaluation approach that relies on multimodal user behaviors.
In this study, our main focus is on assessing the human-likeness of the robot as the primary evaluation metric.
While previous research often relied on subjective evaluations from users, our approach aims to evaluate the robot's human-likeness based on observable user behaviors indirectly, thus enhancing objectivity and reproducibility.
To begin, we created an annotated dataset of human-likeness scores, utilizing user behaviors found in an attentive listening dialogue corpus. 
We then conducted an analysis to determine the correlation between multimodal user behaviors and human-likeness scores, demonstrating the feasibility of our proposed behavior-based evaluation method.
\end{abstract}





\maketitle

\section{Introduction}

One of the research challenges in the field of conversational robots and dialogue systems is the establishment of evaluation methods~\cite{SIM2015305,abd2020technical,deriu2021survey,zhang2021automatic,ni2023survey}.
Thanks to the emergence of large-scale language models (LLMs), recent chatbots have become capable of carrying out highly sophisticated conversations.
The realization of such systems has been made possible by the creation of extensive text datasets and the implementation of evaluation methods.
Since objective evaluation methods alone cannot encompass all phenomena, a series of studies, including human subjective evaluations, have been conducted.
In the case of task-oriented dialogues, such as restaurant searches, since the goal of the conversation is clear and objective, it is straightforward to consider objective evaluation metrics such as task achievement rate and the number of turns, and research and development efforts have been made based on these objective indicators.
However, this is not the case where the target conversations are more real and sophisticated such as ones like human-human conversations in our society.

Recent advancements have led to the development of socially-situated conversational robots (SCRs), which are specifically designed for social contexts.
Socially-situated conversations encompass a wide range of interactions, from brief exchanges such as reception and information guide~\cite{swartout2010ada,iio2020human} to more extended conversations such as counseling~\cite{devault2014simsensei,laranjo2018conversational,scoglio2019use,rasouli2022potential} and interviews~\cite{han-etal-2013-counseling,johnston-etal-2013-spoken,yu2019open,inoue2020job}.
It is crucial to invest efforts into the development of SCRs to enhance their practicality, address diverse social issues, and promote harmonious symbiosis with society.

\begin{figure}[t!]
\centering
\includegraphics[width=60mm]{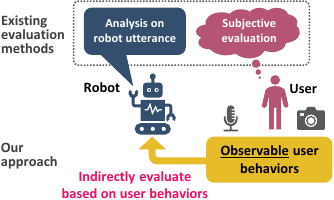}
\caption{Overview of the proposed evaluation scheme}
\label{fig:overview}
\vsapceundergifure
\end{figure}

This study addresses objective evaluation methods for SCRs.
Traditional studies on SCRs have frequently depended on subjective evaluation methods such as ``{\it satisfaction}'' and ``{\it effectiveness}''~\cite{casas2020trends} or actual system utterances due to the lack of a clearly defined goal for the conversation.
However, relying solely on subjective evaluation diminishes research reproducibility and constrains the growth of the research community.
In this study, as an initial step toward the objective and general evaluation method for SCRs, we introduce an evaluation method based on observable and multimodal user behaviors (Figure~\ref{fig:overview}) and report our initial trial in an attentive listening dialogue task.

The contributions of this paper are two folds:
\begin{itemize}
    \item We proposed a new evaluation approach for SCRs based on observable and multimodal user behaviors.
    \item As a target metric, we focused on the human-likeness of the robot and created a dataset together with human-likeness scores annotated based on the user behaviors.
\end{itemize}

\section{Proposed evaluation method}

In the current study, we focus on the concept of {\it human likeness} as a target evaluation metric that may be shared with other studies~\cite{EDLUND2008630,kawahara2019spoken,lin2022duplex}.
The notion of ``{\it naturalness}'' has been adopted as an indicator in numerous studies, and ``{\it human-likeness}'' represents a more concrete manifestation of this concept.
While previous studies have addressed aspects such as user satisfaction~\cite{ultes2021user} and miscommunication~\cite{meena-etal-2015-automatic} in the context of automatic evaluating of conversational systems, the concept of human-likeness pursued in this study differs in its aspiration for natural dialogue between humans, which necessitates more advanced conversational abilities.
SCRs inherently strive for human-to-human social interaction, thus emphasizing the importance of evaluating human likeness rather than naturalness or satisfaction.
Note that the key point of this study is to evaluate SCRs based on multimodal user behaviors, so the proposed evaluation frame can be applied to other evaluation metrics.

We propose an evaluation method that lies in its focus on observable user behaviors.
While conventional evaluation metrics have primarily emphasized system utterances, this has contributed to a heightened subjectivity.
By conducting evaluations based on observable user behavior, we create objective indicators to the fullest extent possible.
User behavior encompasses a wide range of multimodal aspects.
For instance, in addition to including speech and linguistic features such as total utterance time and word count, it encompasses dialogue-specific features such as backchannels, fillers, and switching pause length (turn-taking gap), as well as non-verbal features such as eye gaze, which is specific to embodied conversational robots.

Intuitively, those user behaviors are different depending on the human likeness of the robot.
This can be more understood by comparing conversations between human-robot and human-human ones.
For example, in the context of human-robot conversations, if we contemplate the number of uttered words, the user might tend to utter clearly with a simple and limited vocabulary.
Additionally, in terms of spoken dialogue-specific behaviors, empirical observations indicate that users tend to provide fewer backchannels and have longer turn-taking pauses when interacting with systems that are perceived as non-humanlike.
Conversely, in human-human conversations, they tend to be a proclivity for fluent utterance of a variety of words with smooth turn-taking.
Hence, we can infer that as the number of uttered words increases, proximity to human-human dialogue intensifies, thereby augmenting the human-likeness score.
In this study, we empirically explore multimodal user behaviors that relate to the human likeness of the robot.

\section{Dataset construction}

In this study, we explore the potential of the proposed evaluation framework by utilizing an attentive dialogue corpus.
Here, third-party people subsequently annotated the corpus to give human-likeness scores with a simple approach referring to multimodal user behaviors.

\subsection{Attentive listening dialogue corpus}

An attentive listening dialogue corpus was used in this study.
In this dialogue, the task is to attentively listen to a user's talk, and the system needs to utter the listener responses, such as backchannels and questions.
Several attentive listening systems have been proposed so far~\cite{schroder2015acii,inoue2020sigdial,oertel2021towards}.
In this instance, we employed an existing system~\cite{inoue2020sigdial} for this data collection.
The interface of the system was an android robot~\cite{inoue-etal-2016-talking} whose appearance is similar to that of human beings.
The role of the user was assigned to a university student, who was asked to speak for eight minutes on the topic of ``challenges faced during the COVID-19 pandemic.''
These dialogues were made in the Japanese language.
Note that the aforementioned configuration merely represents one of the potential setups, and it is desirable to explore various types of dialogues, systems, and interfaces in future investigations.

In order to vary the human-likeness of the robot, two scenarios were prepared for this data collection. 
The first scenario entails interacting with the aforementioned pre-existing autonomous system. 
The second scenario involves an operator in a separate room engaging in direct conversation on behalf of the system, the so-called Wizard-of-OZ (WOZ).
In this case, the operator's spoken voice was played back directly through the android's speaker, and nonverbal expressions, such as the android's gaze and gestures, were controlled by the operator using a handheld controller. 
There were a total of two operators, with one of them participating in each dialogue. 
With the two aforementioned configurations, 20 dialogues were recorded using the autonomous system, and 49 dialogues were recorded with operator involvement. 
Thus, there were a total of 69 university students acting as users. 
After each dialogue concluded, the participants were asked to answer a 19-item questionnaire evaluation created in a previous study~\cite{inoue2020sigdial}. 

\begin{figure}[t!]
\centering
\includegraphics[width=65mm]{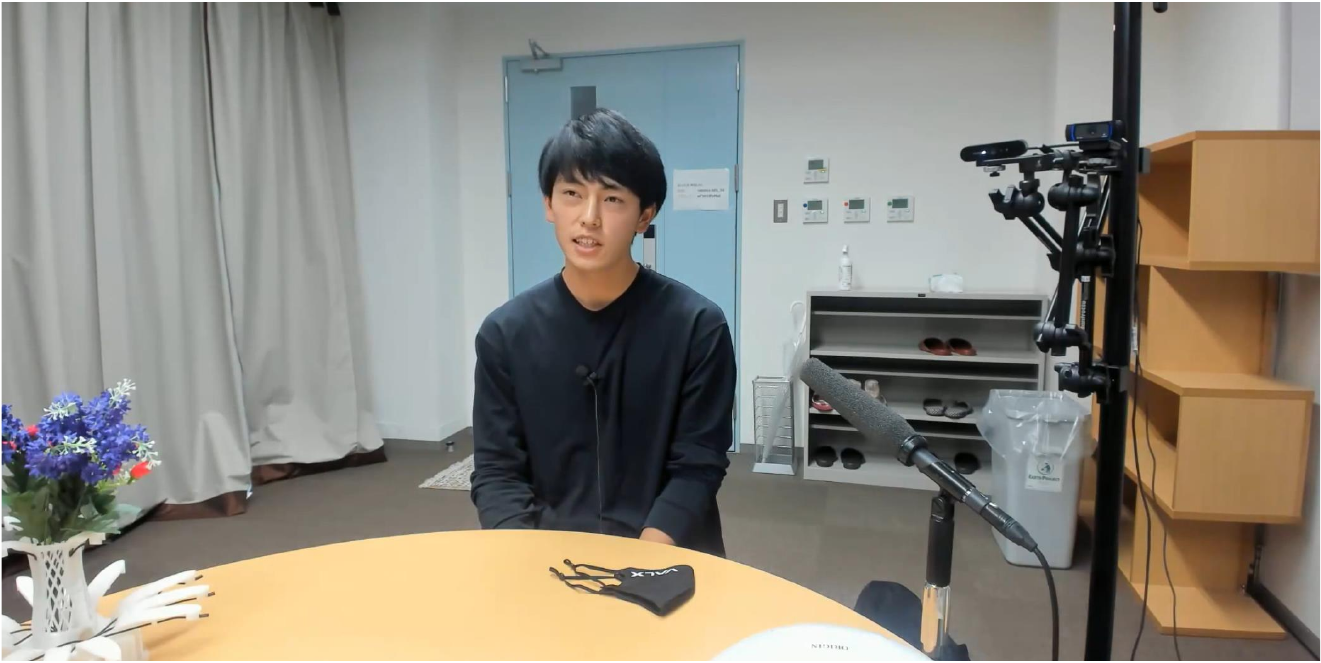}
\vspace{-1mm}
\caption{Sample video clip viewed by annotators}
\label{fig:annotation}
\vsapceundergifure
\end{figure}

\subsection{Annotation of human-likeness scores}

Using the aforementioned dialogue data, we annotated labels to assess the human-likeness of the system.
First, we extracted segments of dialogues and removed the system's visual and auditory components, leaving only the user's visual and auditory inputs.
Third-party annotators were then assigned the task of binary classification to determine whether the dialogue partner (the system) was human or an autonomous system.
Figure~\ref{fig:annotation} illustrates examples of the visual stimuli presented to the annotators.
In other words, the annotators indirectly inferred whether the dialogue partner was a human or a system by focusing solely on the user's behaviors, which are the main focal point of this study.
By gathering judgments from multiple annotators, we calculated the ratio of ``human'' judgments as a measure of ``human-likeness.''
The dialogue segments were extracted for a duration of one minute, resulting in a total of 924 samples from the aforementioned 69 dialogues.
Each sample was assessed by five independent evaluators.
In total, there were 78 annotators, with each being randomly allocated 50 to 70 samples.

Table~\ref{table:annotation_result} presents the results of the annotation.
This evaluation represents the aggregation of sample quantities per numerical value of human-likeness, carried out by the aforementioned annotators.
Initially, when examining the entirety (column ``Total''), it becomes evident that the numerical values of human-likeness exhibit variation.
Next, upon observing the differences between the two system types, it can be discerned that the autonomous system tends to possess comparatively lower proportions of human-likeness.
Meanwhile, WOZ, on the other hand, demonstrates a tendency towards higher numerical values of human-likeness.
In other words, at present, the autonomous system is inferior to WOZ, thus indicating a reasonable reflection of this fact.

\begin{figure}[t!]
\centering
\includegraphics[width=85mm]{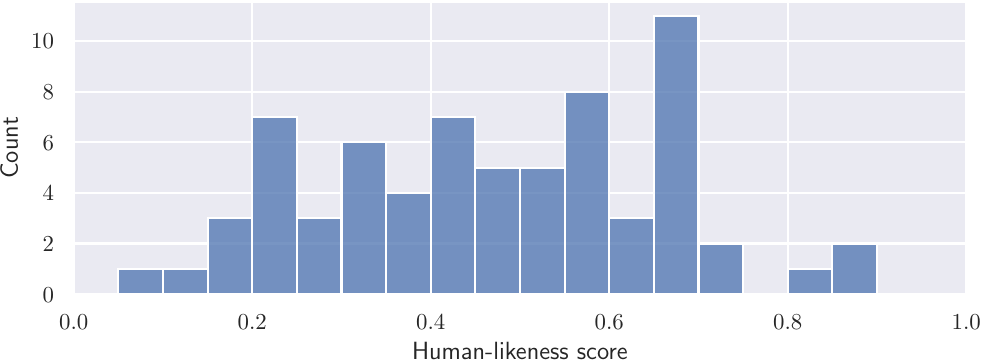}
\caption{Distribution of human-likeness scores averaged per dialogue}
\label{fig:score_average}
\vsapceundergifure
\end{figure}

In this study, to conduct evaluation in each dialogue, we calculated the average score of human-likeness for each dialogue.
The distribution of the averaged scores is illustrated in Figure~\ref{fig:score_average}.
Even in this scenario, it is evident that the scores exhibit variation.
In the subsequent analysis, which is described in the following section, we will use these averaged scores as the target variables.

\begin{table}[t]
  \centering
  \caption{Annotation result of human-likeness score (HL: human-likeness, Auto.: Autonomous system)}
  \vspace{-1mm}
  \label{table:annotation_result}
  \scalebox{0.90}{
\begin{tabular}[width=\hsize]{cccc}
  \hline
  \multirow{2}{*}{HL score} & \multicolumn{2}{c}{System type} & \multirow{2}{*}{Total} \\
  \cline{2-3}
  & Auto. & WOZ & \\
  \hline
  1.0~(5/5) & \phantom{00}4 (\phantom{0}1.5\%) & \phantom{0}66 (10.1\%) & \phantom{0}70 (\phantom{0}7.6\%) \\
  0.8~(4/5) & \phantom{0}26 (\phantom{0}9.7\%) & 138 (21.0\%) & 164 (17.7\%) \\
  0.6~(3/5) & \phantom{0}40 (14.9\%) & 156 (23.8\%) & 196 (21.2\%)  \\
  0.4~(2/5) & \phantom{0}59 (22.0\%) & 145 (22.1\%) & 204 (22.1\%) \\
  0.2~(1/5) & \phantom{0}82 (30.6\%) & 103 (15.7\%) & 185 (20.0\%) \\
  0.0~(0/5) & \phantom{0}57 (21.3\%) & \phantom{0}48 (\phantom{0}7.3\%) & 105 (11.4\%) \\
  \hline
  Total & 268 & 656 & 924 \\
  \hline
  \end{tabular}
  \vsapceundergifure
  }
  \vspace{-3mm}
\end{table}

\section{Analysis}

We then considered the possibility of the proposed evaluation method by investigating the relationship between the annotated human-likeness scores and the multimodal user behaviors.

\subsection{Evaluation of the human-likeness scores from multimodal user behaviors}
\label{sec:analysis:behavior}

\begin{table}[t]
  \centering
  \caption{Correlation coefficients between human-likeness scores and objective user behaviors}
  \vspace{-1mm}
  \label{table:behavior}
  \scalebox{0.90}{
\begin{tabular}[width=\hsize]{ll}
  \hline
  \multicolumn{1}{c}{Behavior} & \multicolumn{1}{c}{Corr. ($r$)} \\
  \hline
  \rowcolor[gray]{0.95} (Voice activity) & \\
  ~~ Total utterance time & {\bf \phantom{-}0.35} \\
  ~~ Average utterance duration & \phantom{-}0.17 \\
  ~~ \# of utterance & \phantom{-}0.11 \\
  \rowcolor[gray]{0.95} (Linguistic) & \\
  ~~ \# of words & \phantom{-}0.16 \\
  ~~ \# of unique words & {\bf \phantom{-}0.33} \\
  ~~ \# of content words & \phantom{-}0.11 \\
  ~~ \# of unique content words & {\bf \phantom{-}0.30} \\
  \rowcolor[gray]{0.95} (Gaze) & \\
  ~~ \# of gaze shift (eye contact) & {\bf \phantom{-}0.21} \\
  ~~ Total gaze duration & \phantom{-}0.00 \\
  ~~ Average gaze duration & -0.12 \\
  \rowcolor[gray]{0.95} (Dialogue) & \\
  ~~ \# of turns & \phantom{-}0.07 \\
  ~~ Average turn duration & \phantom{-}0.04 \\
  ~~ Average switching pause length & {\bf -0.35} \\
  ~~ \# of backchannels & \phantom{-}0.11 \\
  ~~ \# of fillers & \phantom{-}0.05 \\
  ~~ \# of laughs & \phantom{-}0.15 \\
  ~~ \# of disfluencies & -0.03 \\
  \hline
  \end{tabular}
  }
  \vsapceundergifure
\end{table}

The aim of this study is to evaluate the human-likeness scores of SCRs from multimodal user behaviors.
Here, we examined the correlation between the multimodal user behaviors listed in Table~\ref{table:behavior} and the human-likeness scores.
These behaviors can be categorized into four groups: voice activity, linguistic, gaze, and dialogue.
These behaviors are based on manually annotated data, but ones such as voice activity can be extracted automatically.
Content words in the linguistic category are defined as nouns, verbs, adjectives, adverbs, and conjunctions.
The numerical value of each behavior was calculated by averaging those across multiple dialogue segments used in the previous section.

\begin{figure}[t]
\centering
\includegraphics[width=75mm]{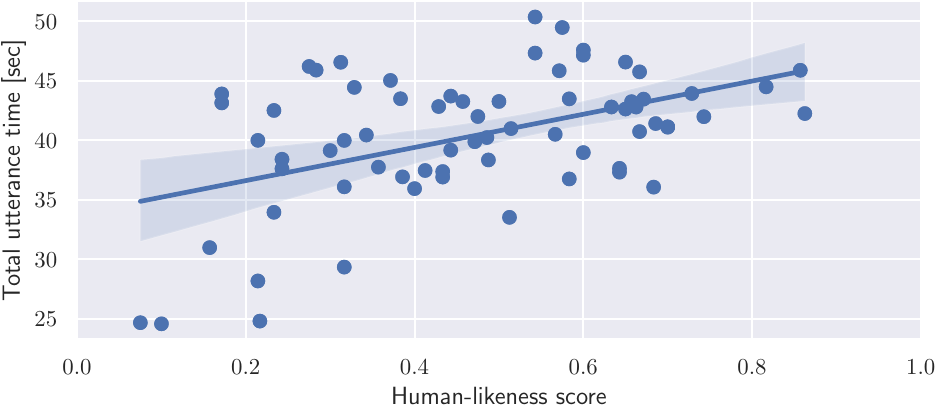} \\
\vspace{2mm}
\includegraphics[width=75mm]{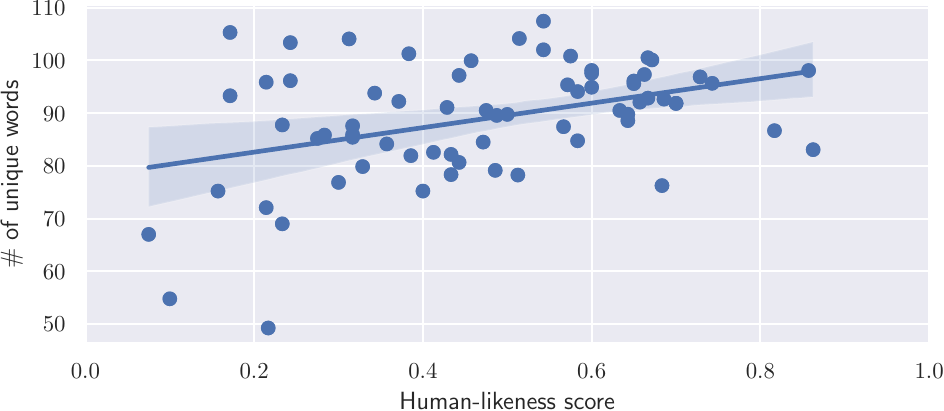} \\
\vspace{2mm}
\includegraphics[width=75mm]{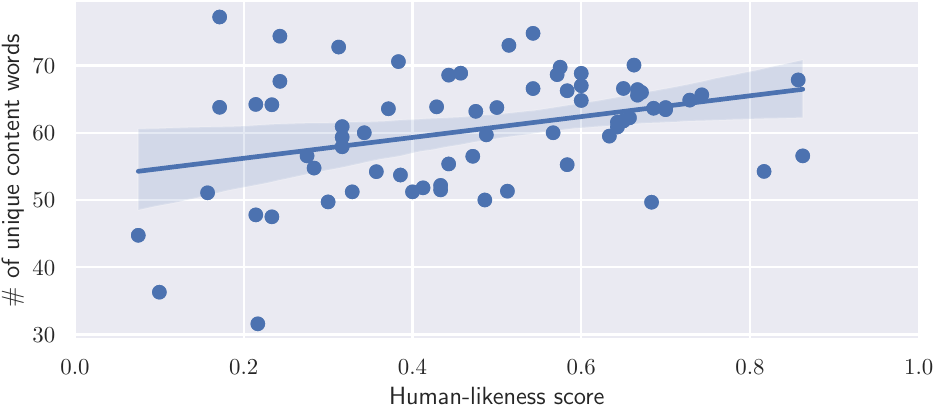} \\
\vspace{2mm}
\includegraphics[width=75mm]{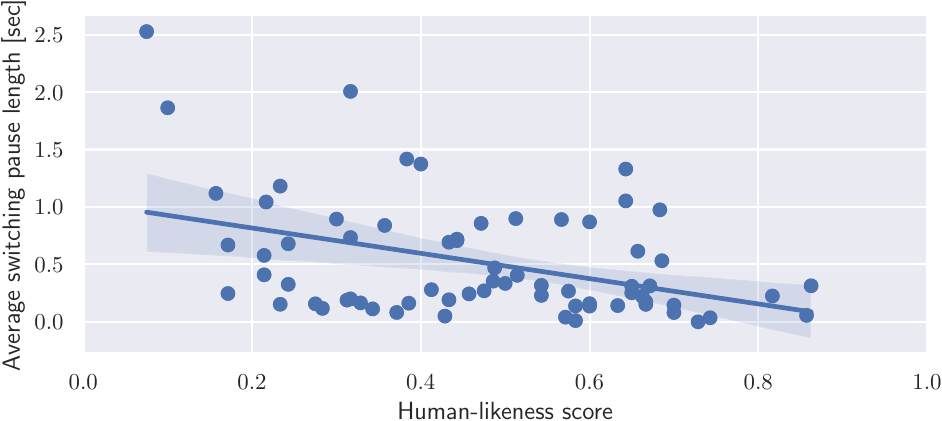}
\caption{Relationship between human-likeness scores and top-4 user behaviors}
\label{fig:appendix:behavior}
\end{figure}

By investigating Spearman's rank correlation coefficients, we observed weak correlations in several behaviors.
Figure~\ref{fig:appendix:behavior} illustrates the correlations on the top-4 user behaviors.
Total utterance time and the number of uttered unique words showed a higher correlation coefficient.
Likewise, the number of gaze shifts and the average switching pause length manifested augmented correlation.
This result indicates the potential of inferring the human-likeness score from these multimodal behaviors.

Therefore, we explored the extent to which the aforementioned behaviors can estimate the human-likeness scores.
We conducted leave-one-out cross-validation using support vector regression.
The target variable was the human-likeness score, and the explanatory variables were the numerical values of the user behaviors listed in Table~\ref{table:behavior}.
The evaluation metric employed was the mean absolute error (MAE).
Consequently, the average MAE amounted to 0.146.
Given that the current dataset consists of values incremented by 0.2, it has been demonstrated that estimating the human-likeness score with an error of less than or equal to one increment is feasible.

\subsection{Relationship with subjective evaluation}

To verify the generalizability and practicality of the human-likeness scores used in this study, we also investigated the relationship with the conventional subjective evaluation scores obtained in the attentive listening dialogue.
Note that the subjective evaluation items were made in the previous study~\cite{inoue2020sigdial}.
Among the 19 evaluation items, there were five items that exhibited weak correlations as listed in Table~\ref{table:subjective}.
For example, the correlated items were ``The robot understood the talk'' ($r=0.39$) and ``I was satisfied with the dialogue'' ($r=0.21$), which are important factors in the attentive listening task.
From these results, the human-likeness scores demonstrate a certain degree of correlation with some subjective evaluations, thereby confirming its generalizability and practicality.

\begin{table}[t]
\setlength{\tabcolsep}{1mm}
  \centering
  \caption{Correlation coefficients between human-likeness labels and subjective evaluation scores}
  \label{table:subjective}
  \scalebox{0.90}{
  \begin{tabular}[width=\hsize]{lp{65mm}c}
  \hline
  \multicolumn{2}{c}{Question item} & Corr. \\
  \hline
  \rowcolor[gray]{0.95} \multicolumn{3}{l}{(Robot behaviors)} \\
  Q1 & The words uttered by the robot were natural &                \phantom{-}0.07 \\
  Q2 & The robot responded with good timing &                       \phantom{-}{\bf 0.25} \\
  Q3 & The robot responded diligently &                             \phantom{-}0.07 \\
  Q4 & The robot's reaction was like a human's &                    \phantom{-}0.19 \\
  Q5 & The robot's reaction adequately encouraged my talk &         \phantom{-}0.07 \\
  Q6 & The frequency of the robot's reaction was adequate &         \phantom{-}0.05 \\[0.2ex]
  \rowcolor[gray]{0.95} \multicolumn{3}{l}{(Impression on the robot)} \\
  Q7 & I want to talk with the robot again &                        \phantom{-}0.08 \\
  Q8 & The robot was easy to talk with &                            \phantom{-}0.12 \\
  Q9 & I felt the robot is kind &                                   \phantom{-}0.10 \\
  Q10 & The robot listened to the talk seriously &                  \phantom{-}0.18 \\
  Q11 & The robot listened to the talk with focus &                 \phantom{-}0.15 \\
  Q12 & The robot listened to the talk actively &                   \phantom{-}0.13 \\
  Q13 & The robot understood the talk &                             \phantom{-}{\bf 0.39} \\
  Q14 & The robot showed interest for the talk &                    \phantom{-}{\bf 0.20} \\
  Q15 & The robot showed empathy towards me &                       \phantom{-} 0.17 \\
  Q16 & I think the robot was being operated by a human &           {\bf -0.30} \\
  Q17 & The robot was good at taking turns &                        \phantom{-}0.19 \\[0.2ex]
  \rowcolor[gray]{0.95} \multicolumn{3}{l}{(Impression on the dialogue)} \\
  Q18 & I was satisfied with the dialogue &                         \phantom{-}{\bf 0.21} \\
  Q19 & The exchange in the dialogue was smooth &                   \phantom{-}0.19 \\
  \hline
  \end{tabular}
  }
\end{table}

\section{Conclusion}

In this paper, we proposed a method to evaluate socially-situated conversational robots based on observable and multimodal user behaviors.
We utilized attentive listening dialogue data for annotation of the human-likeness of the robot, revealing a correlation between the user behaviors and the human-likeness scores.
Additionally, we demonstrated the ability to predict the scores with an average MAE of 0.146.
Future work will involve examining the proposed evaluation method in a wider range of social situations to confirm its generalizability.
For example, we are now extending this work to job interviews and first-time meeting scenarios where the role of the robots is different from the one in the attentive listening scenario.

\section*{Acknowledgement}

This work was supported by JST ACT-X (JPMJAX2103), JST Moonshot R\&D (JPMJPS2011), and JSPS KAKENHI (JP19H05691 and JP23K16901).
The authors also express appreciation to Dr. Masato Komuro for his insightful comments on this research.

\bibliographystyle{ACM-Reference-Format}
\bibliography{reference}

\end{document}